\documentclass[10pt,twocolumn,letterpaper]{article}

\usepackage{cvpr}              
\usepackage{multirow}
\usepackage[table]{xcolor}

\definecolor{cvprblue}{rgb}{0.21,0.49,0.74}
\usepackage[pagebackref,breaklinks,colorlinks,citecolor=cvprblue]{hyperref}
\def\paperID{38} 
\def\confName{CVPR}
\def\confYear{2024}

\title{VeTraSS: Vehicle Trajectory Similarity Search Through
Graph Modeling and Representation Learning}

\author{Ming Cheng$^{1}$, Bowen Zhang$^{2}$, Ziyu Wang$^{3}$, Ziyi Zhou$^{1}$, Weiqi Feng$^{4}$, Yi Lyu$^{5}$, Xingjian Diao$^{1}$ \\
$^1$Dartmouth College
    $^2$Shanghai Jiao Tong University 
    $^3$University of California, Irvine \\
    $^4$Harvard University $^5$Independent Researcher \\
{\tt\small \{ming.cheng.gr, ziyi.zhou.gr, xingjian.diao.gr\}@dartmouth.edu}\\
{\tt\small zzljoy3091498@sjtu.edu.cn ziyuw31@uci.edu}\\
{\tt\small wfeng@g.harvard.edu isabellalyu1130@gmail.com}
}

\begin{document}
\maketitle
\begin{abstract}
Trajectory similarity search plays an essential role in autonomous driving, as it enables vehicles to analyze the information and characteristics of different trajectories to make informed decisions and navigate safely in dynamic environments. 
Existing work on the trajectory similarity search task primarily utilizes sequence-processing algorithms or Recurrent Neural Networks (RNNs), which suffer from the inevitable issues of complicated architecture and heavy training costs.
Considering the intricate connections between trajectories, using Graph Neural Networks (GNNs) for data modeling is feasible.
However, most methods directly use existing mathematical graph structures as the input instead of constructing specific graphs from certain vehicle trajectory data. This ignores such data's unique and dynamic characteristics.
To bridge such a research gap, we propose VeTraSS -- an end-to-end pipeline for \textbf{Ve}hicle \textbf{Tra}jectory \textbf{S}imilarity \textbf{S}earch. 
Specifically, VeTraSS models the original trajectory data into multi-scale graphs, and generates comprehensive embeddings through a novel multi-layer attention-based GNN. 
The learned embeddings can be used for searching similar vehicle trajectories. 
Extensive experiments on the Porto and Geolife datasets demonstrate the effectiveness of VeTraSS, where our model outperforms existing work and reaches the state-of-the-art.
This demonstrates the potential of VeTraSS for trajectory analysis and safe navigation in self-driving vehicles in the real world. 
\end{abstract}    
\section{Introduction}
The advancement of autonomous driving technologies has fundamentally changed transportation, leveraging cloud computing \cite{yao2020privacy, zhang2021tapping, zhang2023first} and Location-Based Services (LBS) \cite{hu2020dasgil, shaheer2023graph, kabalar2023towards} to interpret the complex patterns of vehicular movements and human activities. This shift requires analyzing vast spatio-temporal data for improved traffic safety \cite{yang2022zebra} and management \cite{narmadha2023spatio}, despite challenges from its volume and diversity. Data analysis innovations are crucial for integrating autonomous vehicles smoothly into our lives. This data offers significant potential for intelligent decision-making in health monitoring \cite{zhou2021doseguide, cheng2023saic, wang2020guardhealth, wang2024differential} and mobile computing \cite{kwon2023neural, alikhani2024seal}, enhancing efficiency and safety in our interconnected world.

\begin{figure}[tb]
  \centering
  \includegraphics[width=0.8\linewidth]{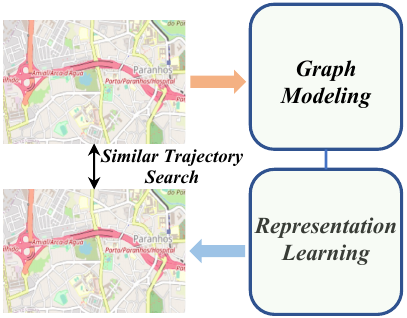}
  \caption{\textbf{Trajectory similarity search pipeline.} 
    VeTraSS constructs the graph from the original vehicle trajectory data, followed by graph representation learning and embedding generation. The embedding is used for accurately searching alternative trajectories. 
  }
  \vspace{-15pt}
  \label{fig:teaser}
\end{figure}

Traditionally, dimensionality reduction algorithms have served to distill the complexity of trajectory data into lower-dimensional spaces, aiding in the analysis of vehicle trajectories for autonomous driving \cite{wold1987principal, kruskal1978multidimensional, bjornsson1997manual}. These methods, while providing a foundation for understanding vehicular patterns, encounter limitations, including slow inference speeds and challenges in accurately representing the intricate details of diverse and novel real-world scenarios \cite{bjornsson1997manual, bashir2005hmm, richards1995trajectory}.

In autonomous driving, accurately predicting vehicular movement via trajectory analysis is crucial, requiring sophisticated models to manage the complexity of vehicle trajectories. Recent shifts towards neural network-based approaches, especially those leveraging sequence processing capabilities like LSTM and RNN, have marked significant progress in spatio-temporal representation learning \cite{li2018deep, yao2019computing, zhang2020trajectory, zhang2019deep, huang2023lstm}. Yet, these models frequently entail extensive data pre-processing and bear the brunt of complex, resource-intensive training processes.

Graph Neural Networks (GNNs) \cite{qiu2020gcc, liu2020dynamic, xu2018powerful, velivckovic2018deep} have shown promise in spatio-temporal data analysis with their graph-based modeling. However, their application in autonomous driving, especially for vehicle trajectory analysis, remains underexplored. The common use of generic graph structures instead of creating specific graphs from vehicle trajectory data fails to capture its unique dynamics \cite{tang2015line, perozzi2014deepwalk, grover2016node2vec}. This gap prompts our research question: \textbf{\textit{How can we design an end-to-end pipeline that creates tailored graph structures from vehicle trajectory data, precisely capturing autonomous driving dynamics to improve the accuracy and relevance of trajectory similarity search?}}

To bridge this research gap, we present VeTraSS, an innovative end-to-end pipeline for vehicle trajectory similarity search, as shown in Figure \ref{fig:teaser}. VeTraSS models vehicle trajectories using multi-scale graphs and employs a multi-layer attention-based GNN to produce detailed embeddings to improve the search for similar trajectories, aiding autonomous vehicles in making safe, informed decisions. 

In summary, our contribution is threefold:
\begin{itemize}
    \item {
    \textbf{End-to-end pipeline for vehicle trajectory analysis.}
    To the best of our knowledge, VeTraSS is the \textit{\textbf{first}} pipeline tailored specifically for the autonomous driving sector, handling the entire process from graph construction of vehicle trajectories to the generation of their representations. 
    }
    \item {
    \textbf{State-of-the-art performance. }
    Our rigorous experiments on the Porto \cite{moreira2016time} and Geolife \cite{zheng2010geolife} datasets, both integral to autonomous driving research, validates VeTraSS's capability. It outperforms existing methodologies in accuracy and sets a new benchmark in vehicle trajectory similarity search.
    }
    \item {
    \textbf{Time-efficiency for representation learning. }
    Different from LSTM/RNN-based models, VeTraSS leverages an attention mechanism for multi-scale embedding generation. Experimental results confirm that this design excels in real-time autonomous driving applications, where quick decision-making is essential.
    }
\end{itemize}

\begin{figure*}[tb]
  \centering
  \includegraphics[width=0.9\linewidth]{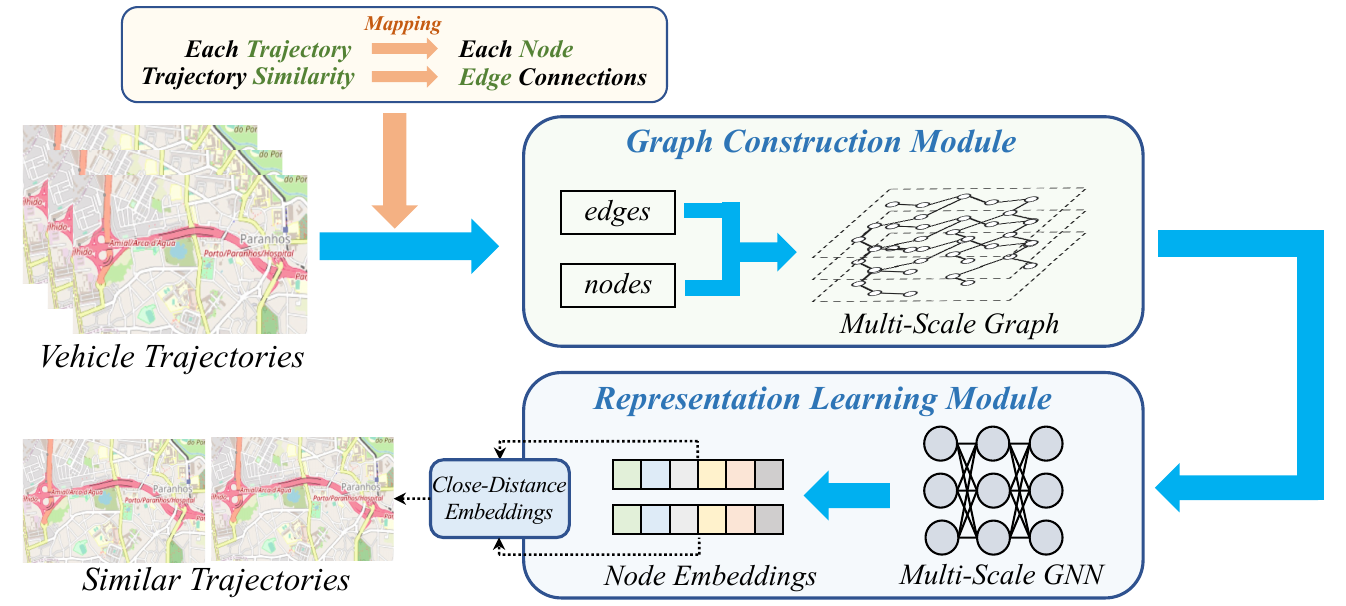}
  \caption{\textbf{Overview of the VeTraSS pipeline.}
  \textbf{Graph construction and embedding generation:} 
  The original high-dimensional trajectories are mapped into low-dimensional space for graph construction, where each node corresponds to each trajectory while edge connections represent trajectory similarity degrees. The multi-scale graph is input into a multi-scale GNN to generate accurate node embeddings. 
  \textbf{Trajectory similarity search:} Given a query trajectory, VeTraSS generates the corresponding embedding vector and also finds the closest embedding vector to it. The closest embedding represents the most similar trajectory.
  }
  \label{fig:pipeline}
\end{figure*}

\section{Related Work}

\subsection{Non-Learning-Based Methods}

In autonomous driving, traditional dimensionality reduction methods like PCA \cite{wold1987principal}, SVD \cite{bjornsson1997manual}, and MDS \cite{chen2020graph} were initially used for learning representations of vehicle trajectory data. These techniques aimed to reduce the complexity of spatio-temporal data by mapping it to a lower-dimensional space, making it easier to analyze and interpret. PCA works by maximizing the variance of projected sample data on a hyperplane, enhancing data representation. SVD creates a low-dimensional space through matrix decomposition, and MDS focuses on maintaining sample distances after reduction.

Despite the mathematical rigor and the theoretical promise of these techniques, they exhibit certain limitations when applied to the dynamic and varied datasets typical of autonomous driving scenarios. Specifically, they tend to suffer from slow inference speeds and a lack of flexibility in handling the diverse and unpredictable nature of real-world driving data \cite{lim2019hybrid, gu2013focused}. This gap between theoretical effectiveness and practical applicability has prompted the exploration of more advanced methods capable of addressing the unique challenges posed by autonomous driving data.

\subsection{Sequence-Processing-Based Methods}

To overcome the limitations of non-learning methods, deep learning models employing sequence processing strategies have been proposed. Sukhbaatar et al. \cite{sukhbaatar2015end} introduced an RNN structure for capturing long-term dependencies, which Chandar et al. \cite{chandar2016hierarchical} extended to a hierarchical architecture. Moreover, approaches like SRN \cite{pei2016modeling} and NEUTRAJ \cite{yao2019computing} use a combination of memory networks (LSTM + RNN) for trajectory analysis, with \cite{zhou2023grlstm} integrating Residual-LSTM for enhanced representation learning. Despite their ability to capture time-dependent features, the complexity of recurrent architectures demands significant training resources, limiting their practical application in autonomous driving.

\subsection{GNN-Based Methods}

Addressing the shortcomings of previous models, GNN-based approaches have emerged for graph representation learning. GGS-NN \cite{li2015gated} employs gated recurrent units in a GNN framework, while node2vec \cite{grover2016node2vec} innovates in defining node neighborhoods and random walk procedures for feature representation. Liu et al. \cite{liu2023beyond} introduced an unsupervised model that distinguishes between homophilic and heterophilic edges. However, these GNN models, designed for general graph-structured data, often overlook the specific requirements of spatio-temporal data in autonomous driving, such as capturing temporal dynamics and movements.

Building on this foundation, we introduce VeTraSS, a novel end-to-end pipeline that unifies spatio-temporal-specific graph construction with comprehensive representation learning, targeting the intricate dynamics of vehicle trajectories in autonomous driving environments.

\section{Method}
Considering the impossibility of directly conducting the similarity search task from the original dataset, the mathematical abstraction that maps from a high-dimensional space (trajectory dataset) into a low-dimensional one (node embedding in graphs) is necessary. 
 After constructing the graph, a novel GNN model is designed to learn graph representations and generate node embeddings.
The low-dimensional embedding vector corresponds to each high-dimensional trajectory. 

The overview of the VeTraSS pipeline is shown in Figure \ref{fig:pipeline}.
VeTraSS mainly consists of two parts, with the first one constructing the multi-scale graph from the original dataset, and the second one focusing on designing a novel attention-based GNN for precise representation learning. 

\subsection{Graph Construction Module}
\label{sec:graphcons}

\subsubsection{Similarity Matrix Generation} 
\label{simigen}
Assume the original vehicle trajectory dataset as $\mathcal{T} = \{T_1, T_2, ..., T_n\}$ where $T_i$ indicates each trajectory containing $l$ data points:
\begin{equation}
\label{eq_trace}
    T_i = \{(P_1, t_1), (P_2, t_2), ..., (P_l, t_l)\}
\end{equation}
Here, $P_k \in \mathcal{R}^2$ is the spatial position point corresponding to the sampling time $t_k$, which generally contains information on longitude and latitude. 
The relative magnitude of the distance between samples is used to model the similarity of two trajectories. We define the normalized distance $d_{ij}$ between each sample under certain distance function $dist(\cdot, \cdot)$ as:
\begin{equation}
\label{eq1}
    d_{ij} = 1 - \frac{e^{-dist(T_i, T_j)}}{\sum_{k=1}^n e^{-dist(T_i, T_k)}}
\end{equation}
where $dist(\cdot, \cdot)$ is Fréchet distance \cite{frechet1906quelques} / Hausdorff distance \cite{belogay1997calculating}.
The distance matrix $M = (d_{ij})_{n\times n}$ is therefore constructed for the multi-layer similarity graph generation afterward. 

\subsubsection{Multi-Scale Graph Construction}
To allow graph neural networks (GNNs) to process and model the original dataset properly, each trajectory in the dataset is mapped as a node in the graph, and the edges indicate the connection weights between two nodes, as inspired by \cite{yao2019computing}.
Assume the multi-scale undirected weighted graph $\mathcal{G}$ has $m$ layers in total, it can be expressed as: 
\begin{equation}
    \mathcal{G} = \{G_1(V, E_1, A_1), G_2(V, E_2, A_2), ..., G_m(V, E_m, A_m)\}
\end{equation}
where $G_k(V, E_k, A_k)$ represents the $k^{th}$ layer of the graph, and $V$ is the set of nodes corresponding to the trajectories in the dataset. $E_k$ and $A_k = (a_{k, i, j})_{n\times n}$ indicates the edges and the adjacent matrix of the $k^{th}$ layer graph, respectively. 

Since each node represents each trajectory, the node connection situations are determined by the similarity distance of trajectories (Equation \ref{eq1}). 
Specifically, since too close or distant nodes cannot establish a connection, a threshold value $c_i$ (both upper and lower bounds) is involved in determining whether there is a connection between node $v_i$ and $v_j$ in each layer of the graph.
Formally, 
$a_{k, i, j}$, as the adjacent item between node $v_i$ and $v_j$ in layer $k$, can be defined as:
\begin{equation}
\label{eq_a}
    a_{k, i, j} = 
    \begin{cases}
      1 - d_{ij} & \text{if}\quad c_{k-1} \leq d_{ij} \leq c_k \\
      0 & \text{otherwise}
    \end{cases}   
\end{equation}

Therefore, if $a_{k, i, j} >0$, we can assume $v_i$ and $v_j$ form a connection in layer $k$, and consequently, trajectory $T_i$ and $T_j$ have a degree of similarity in that layer.

\subsubsection{Threshold Determination}
The threshold value mentioned previously needs to be determined properly. 
Since the distance situation satisfies the trigonometric inequality, which means for arbitrary trajectory $T_r, T_p, T_q$ in the data $\mathcal{T}$, we must have:
\begin{equation}
    d_{rp} + d_{rq} \geq d_{pq}
\end{equation}
Without loss of generality, 
if assuming  
nodes $(v_p, v_r)$ and $(v_q, v_r)$ are neighbor nodes, $(v_p, v_q)$ must not be. Therefore, following Equation \ref{eq_a}, we have:
\begin{equation}
    \begin{cases}
      c_{k-1} \leq d_{rp} < c_k \\
      c_{k-1} \leq d_{rq} < c_k \\
      d_{pq} \geq c_k
    \end{cases}
\end{equation}
Moreover, if we need to build the connection between $(v_p, v_q)$ in $k+1$ layer to let the graph represent the node connection, we have:
\begin{equation}
    c_k \leq d_{pq} <2c_{k}
\end{equation}
Therefore, if we choose the threshold value as:
\begin{equation}
    c_k = 2c_{k-1}
\end{equation}
we can guarantee that $(v_p, v_q)$ in $k+1$ layer will form a connection, showing the comprehensive graph representation of the original data.

Following the steps above, the connection situation between each node in each layer can be determined, and therefore, the overall graph with $m$ layers is constructed. 

\begin{figure}[tb]
  \centering
  \includegraphics[width=\linewidth]{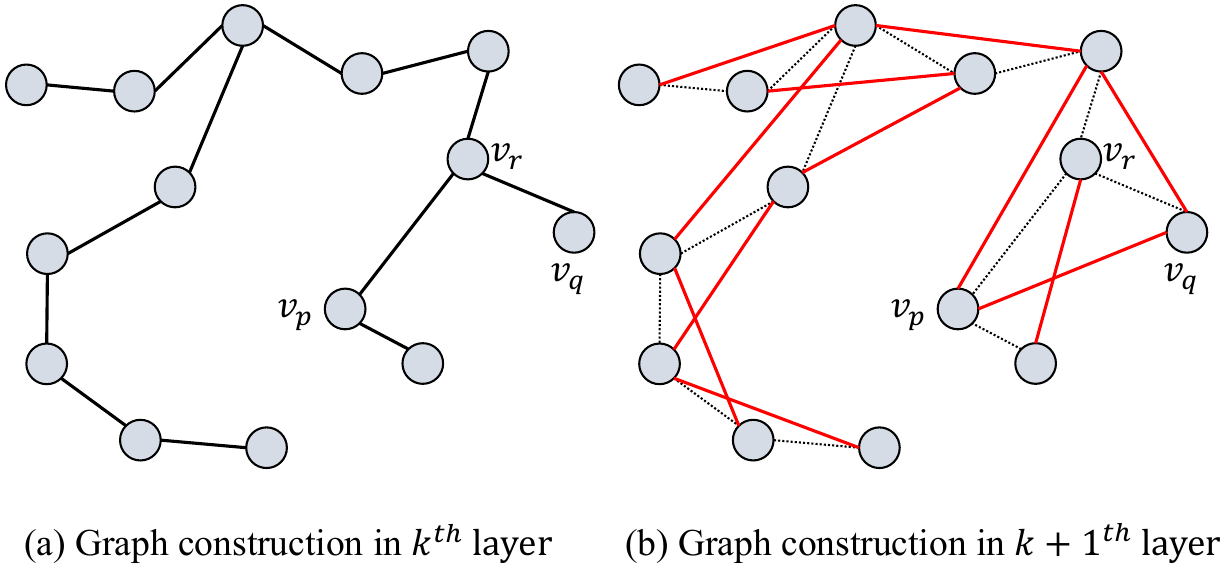}
  \caption{\textbf{Multi-scale graph construction.} 
  Edge between node $v_r$ and $v_p$ shows the similarity between trajectory $\mathcal{T}_r$ and $\mathcal{T}_p$.
  The nodes that are not connected in the $k^{th}$ layer (with a distance of 2) will form a connection in the $k+1^{th}$ layer, replacing the original edges. 
  }
  \label{fig:graph}
\end{figure}

The qualitative example of the algorithm is shown in Figure \ref{fig:graph}
, where red edges indicate the newly established connection formed in the $k+1$ layer between nodes with a distance of 2 in layer $k$.

\begin{figure}[tb]
  \centering
  \includegraphics[width=\linewidth]{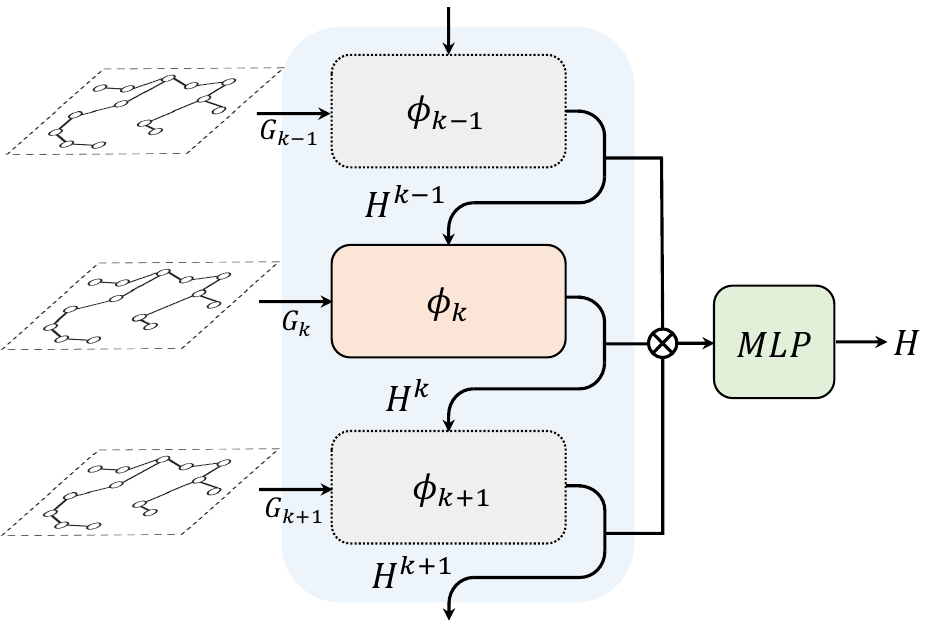}
  \caption{\textbf{Graph representation learning module.} 
  The figure illustrates the $k^{th}$ layer of the model, where the inputs are the graph of layer $k$ ($G_k$) and the embedding of layer $k-1$ ($H^{k-1}$), and outputs the embedding of the $k^{th}$ layer ($H^{k}$). The final output of the model is the concatenation of each layer's output followed by an MLP module.
  }
  \label{fig:arch}
\end{figure}

\subsection{Representation Learning Module}

Through Section \ref{sec:graphcons}, the graph with multi-scale is constructed, which can be further input into GNN for embedding representation learning. 

The architecture design is shown in Figure \ref{fig:arch},
representing an $m$-layer multi-scale GNN ($\phi$).
Since the entire model is designed in a sequential connected pattern, the module ($\phi_k$) in layer $k$ takes the $k^{th}$ layer graph ($G_k$) and the $k-1^{th}$ layer embedding representation ($H^{k-1}$) as the input to generate the $k^{th}$ layer one.
Specifically, 
\begin{equation}
\label{equa_h}
    H^{k-1} = (h_1^{k-1}, h_2^{k-1}, ..., h_n^{k-1})
\end{equation}
where $h_i^{k-1} \in \mathcal{R}^{1\times D_{k-1}}$ represents the embedding vector of node $v_i$ in layer $k-1$, and $D_{k-1}$ is the dimension of the vector.

\subsubsection{Attention Mechanism}
Inspired by existing attention-based architectures \cite{vaswani2017attention, diao2023ft2tf, dosovitskiy2020image, diao2023av}, an attention mechanism is employed for each layer of $\phi$ to measure the influence degree between the nodes. 
Formally, the attention coefficient between node $i$ and $j$ in layer $k$ can be expressed as below:
\begin{equation}
\label{att}
    w_{k,i,j} = attn(W_kh_i^{k-1}, W_kh_j^{k-1})
\end{equation}
where $W_k$ represents a trainable parameter.

To avoid the paradigm expansion of the hidden state vectors which leads to the gradient explosion during training, the attention coefficient is further normalized as below:
\begin{equation}
\label{noratt}
    \alpha_{k,i,j} = \frac{e^{w_{k,i,j}}}{\sum_{v_p \in \mathcal{N}_k(v_i)} e^{w_{k,i,p}}}
\end{equation}
where $\mathcal{N}_k(v_i)$ indicates the neighbor node of $v_i$ in layer $k$.

\begin{table*}[htbp]
\centering
\resizebox{0.9\textwidth}{!}{
\begin{tabular}{@{}c|c|c|ccc|ccc@{}}
\toprule
                                   &                                    &                                     & \multicolumn{3}{c|}{\textbf{Fréchet}}                                                                                       & \multicolumn{3}{c}{\textbf{Hausdorff}}                                                                                      \\ \cmidrule(l){4-9} 
\multirow{-2}{*}{\textbf{Methods}} & \multirow{-2}{*}{\textbf{Dataset}} & \multirow{-2}{*}{\textbf{Backbone}} & \textbf{HR@10}                          & \textbf{HR@50}                          & \textbf{R10@50}                         & \textbf{HR@10}                          & \textbf{HR@50}                          & \textbf{R10@50}                         \\ \midrule
PCA          \cite{wold1987principal}                      &                                    & Non-learning                        & 0.4203                                  & 0.4909                                  & 0.8038                                  & 0.4850                                  & 0.5439                                  & 0.8454                                  \\
SVD       \cite{wall2003singular}                         &                                    & Non-learning                        & 0.4294                                  & 0.4977                                  & 0.8106                                  & 0.4839                                  & 0.5436                                  & 0.8445                                  \\
MDS          \cite{kruskal1978multidimensional}                      &                                    & Non-learning                        & 0.4661                                  & \textbf{0.5874}                         & 0.8607                                  & 0.4839                                  & 0.6065                                  & 0.8770                                  \\ \cmidrule(r){1-1} \cmidrule(l){3-9} 
SRN        \cite{pei2016modeling}                        &                                    & RNN+LSTM                            & 0.4720                                  & 0.5828                                  & 0.7749                                  & 0.3800                                  & 0.4998                                  & 0.7421                                  \\
NEUTRAJ    \cite{yao2019computing}                        &                                    & RNN+LSTM                            & 0.4801                                  & 0.5627                                  & 0.8266                                  & 0.5025                                  & 0.6051                                  & 0.8492                                  \\
node2vec       \cite{grover2016node2vec}                    &                                    & GNN                                 & 0.4175                                  & 0.4519                                  & 0.7376                                  & 0.4317                                  & 0.4637                                  & 0.7602                                  \\
\textbf{VeTraSS (Ours)}            & \multirow{-7}{*}{Porto}            & GNN                                 & \cellcolor[HTML]{DAE8FC}\textbf{0.4968} & \cellcolor[HTML]{DAE8FC}0.5853          & \cellcolor[HTML]{DAE8FC}\textbf{0.8841} & \cellcolor[HTML]{DAE8FC}\textbf{0.5132} & \cellcolor[HTML]{DAE8FC}\textbf{0.6275} & \cellcolor[HTML]{DAE8FC}\textbf{0.8953} \\ \midrule
PCA   \cite{wold1987principal}                             &                                    & Non-learning                        & {0.4336}                         & 0.5880                                  & 0.8446                                  & 0.4110                                  & 0.5562                                  & 0.8243                                  \\
SVD      \cite{wall2003singular}                          &                                    & Non-learning                        & 0.4438                                  & 0.6041                                  & 0.8448                                  & 0.4081                                  & 0.5563                                  & 0.8248                                  \\
MDS            \cite{kruskal1978multidimensional}                    &                                    & Non-learning                        & 0.4793                                  & 0.6187                                  & 0.8716                                  & 0.3602                                  & 0.5472                                  & \textbf{0.8535}                                  \\ \cmidrule(r){1-1} \cmidrule(l){3-9} 
SRN           \cite{pei2016modeling}                     &                                    & RNN+LSTM                            & 0.4631                                  & 0.6032                                  & 0.8121                                  & 0.3120                                  & 0.4236                                  & 0.6640                                  \\
NEUTRAJ     \cite{yao2019computing}                       &                                    & RNN+LSTM                            & 0.4947                                  & \cellcolor[HTML]{DAE8FC}\textbf{0.6786}                                  & 0.8403                                  & 0.3691                                  & 0.4870                                  & 0.7416                                  \\
node2vec      \cite{grover2016node2vec}                     &                                    & GNN                                 & 0.4064                                  & 0.4565                                  & 0.7268                                  & 0.3956                                  & 0.4537                                  & 0.7461                                  \\
\textbf{VeTraSS (Ours)}            & \multirow{-7}{*}{Geolife}          & GNN                                 & \cellcolor[HTML]{DAE8FC}\textbf{0.5003}          & {0.6765} & \cellcolor[HTML]{DAE8FC}{\textbf{0.8831}} & \cellcolor[HTML]{DAE8FC}\textbf{0.4862} & \cellcolor[HTML]{DAE8FC}\textbf{0.6459} & \cellcolor[HTML]{DAE8FC}\textbf{0.8535} \\ \bottomrule
\end{tabular}
}
\caption{
\textbf{
Quantitative comparisons with state-of-the-art models.}
VeTraSS outperforms existing state-of-the-art models on the Porto and Geolife datasets under Fréchet and Hausdorff distance.
Numbers in \textbf{bold}: Highest among all methods. Numbers in \colorbox[HTML]{DAE8FC}{blue} background: Highest among learning-based methods.
}
\label{bigtable}
\end{table*}

\subsubsection{Embedding Representation Generation}
The embedding representation of each node $v_i$, denoted as $h_i^k \in \mathcal{R}^{1\times D_k}$ ($D_k$ is the dimension), can be expressed through an activation function $\sigma(\cdot)$ and the normalized attention coefficient in Equation \ref{noratt}:
\begin{equation}
    h_i^k = \sigma(\sum_{v_j \in \mathcal{N}_k(v_i)}\alpha_{k,i,j}W_kh_j^{k-1})
\end{equation}
Afterward, as expressed in Equation  \ref{equa_h}, the final output of the model in layer $k$ is the collection of all $n$ nodes:
\begin{equation}
    H^k = (h_1^k, h_2^k, ..., h_n^k)
\end{equation}
where $h^k_i$ indicates the embedding of node $i$ (corresponds to trajectory $T_i$) in layer $k$.

\subsubsection{Model Output}
Through all $m$ layers, the final output of the model $\phi$ is the concatenation of all $k$ layers:
\begin{equation}
    H = MLP(Con(H^1, H^2, ..., H^k))
\end{equation}
where $Con$ represents the concatenation operation and $H = (h_1, h_2, ...h_n)$ indicates a set of $n$ embeddings generated by the model, and each embedding corresponds to a trajectory in the dataset. 

Therefore, the original high-dimensional trajectory dataset ($T_i \in \mathcal{T}$) can be mathematically expressed as a low-dimensional embedding vector ($h_i \in H$), which can be used for efficient trajectory similarity search. 
Detailed procedures of trajectory similarity search are introduced in Section \ref{eval}.

We employ cosine similarity distance as the loss function $L$ for model training, where the loss is constructed between the generated embeddings $H \in \mathcal{R}^{N\times D}$ ($N, D$ represent the number of trajectories and embedding dimensions, respectively) and the ground truth similarity distances $M \in \mathcal{R}^{N\times N}$ (Section \ref{simigen}):
\begin{equation}
    L = Cosine(HH^T, M)
\end{equation}
to instruct the model to generate accurate embeddings that are close to the actual similarity matrix. 
\section{Experiments}
\subsection{Dataset}
We conduct extensive experiments on two widely used large-scale datasets collected from the real world, Porto \cite{moreira2016time} and Geolife \cite{zheng2010geolife}, to thoroughly evaluate VeTraSS's performance. 
The Porto dataset captures over 1.7 million taxi trajectories located in the Porto area in Portugal. Specifically, it contains 1,704,759 trajectory counts, with an average of 60 data points per trajectory. The collected data are in the longitude of $(-8.735152,-8.156309)$ and the latitude of $(40.953673,41.307945)$. 
Meanwhile, the Geolife dataset by Zheng \textit{et al.} \cite{zheng2010geolife} presents a comprehensive collection of GPS trajectories that contains the movements of 182 individuals over five years. Specifically, it comprises over 24,876 trajectories, covering a distance exceeding 1.2 million kilometers and totaling more than 48,000 hours in duration. Geographically, the data encompasses multiple cities in China, spanning longitudes from $115.9$ to $117.1$ and latitudes from $39.6$ to $40.7$. 

As publicly available real-world trajectory datasets, Porto and Geolife well reflect the complexity and variability of real-world traffic trajectories, making it an ideal dataset to be used for model performance evaluation.

\begin{figure*}[tb]
  \centering
  \includegraphics[width=0.9\textwidth]{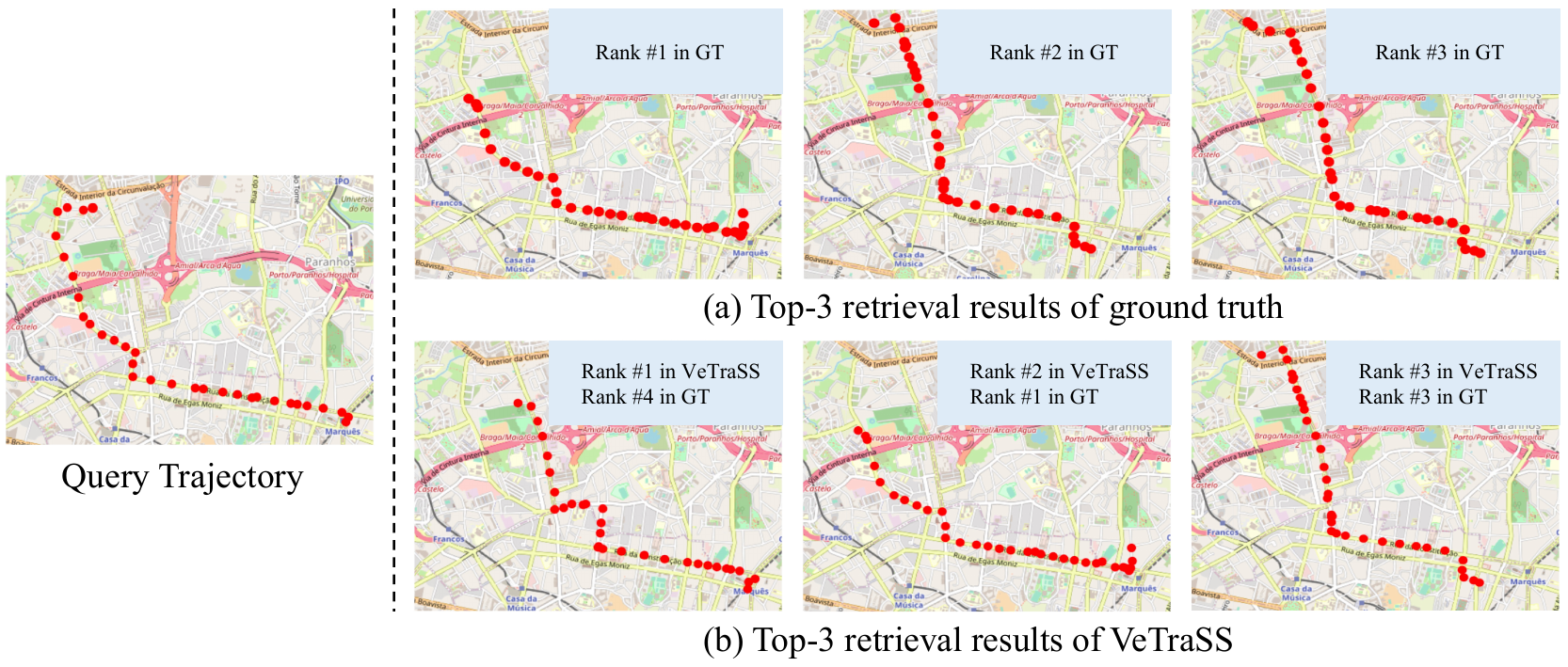}
  \caption{\textbf{Qualitative visualization.} 
  \textbf{Left:} Query trajectory. The three most similar trajectories are retrieved on the right. 
  \textbf{Right (a): } Top-3 retrieval results of ground truth by computing the real distance as mentioned in Section \ref{eval}. 
  \textbf{Right (b): } Top-3 retrieval results of VeTraSS. It is observed that the retrieval results of VeTraSS closely match the ground truth, qualitatively proving its effectiveness.
  }
  \label{fig:qualitativeres}
\end{figure*}

\subsection{Implementation Details}
We preprocess the dataset by removing the trajectories having less than 50 data points, and divide the dataset into 50m$\times$50m grids, following \cite{yao2019computing}. 
The embedding dimension in the model's middle layer is set as $256$, while the final output embedding dimension is $128$. The activation function chosen for embedding generation is ReLU. We use StepLR as the learning rate scheduler, which decreases by a factor of 0.1 every 5 epochs. 
The initial embedding vector is generated through Gaussian random initialization.
The experiments are conducted on the Intel Xeon Silver 4116 @ 2.10GHz CPU and NVIDIA RTX 2080Ti GPU. The model's parameters are updated through SGD (Stochastic Gradient Descent) and Adam optimizer. 

\subsection{Evaluation Procedure}
\label{eval}
We conduct the \textit{top-N} similarity search task to evaluate the model's performance: Given a query trajectory, the model finds $N$ trajectories that are most similar to it.

Assume $\mathcal{T}$ is the set of all trajectories in the dataset, we follow the steps below:
\begin{enumerate}
    \item {    
    For each trajectory $T \in \mathcal{T}$, a specific similarity distance measurement (Fréchet \cite{frechet1906quelques} / Hausdorff distance \cite{belogay1997calculating}) is used to compute the ground truth distance between $T$ and the remaining ones to select its \textit{top-N} similar trajectories. These $N$ trajectories from the set $Y_T$ are expressed below:
    \begin{equation}
        Y_T = \{T^G_1, T^G_2, ..., T^G_N\}, T^G_i \in \mathcal{T}
    \end{equation}
    where $G$ indicates the ground truth domain while $Y_T$ is the \textit{top-N} similar trajectory set for trajectory $T$ specifically.
    }
    \item {
    For each trajectory $T$, the model outputs the embedding representation  $h$ correspondingly. Therefore, the representation of all $T \in \mathcal{T}$ can be expressed as:
    \begin{equation}
        H = \{h_1, h_2, ..., h_M\}
    \end{equation}
    where $M$ indicates the total number of trajectories. 
    }
    \item {
    For each $h \in H$, the Euclidean distance is computed between $h$ and the remaining ones to form the distance matrix of all representations: $d_{ij} = ||h_i - h_j||_2$.
    Therefore, for each trajectory $T$, the \textit{top-K} similar trajectories set ($X_T$) based on the above representation distance can be formed by ranking $d_{ij}$:
    \begin{equation}
        X_T = \{T^P_1, T^P_2, ..., T^P_K\}, T^P_i \in \mathcal{T}
    \end{equation}
    }
    where $P$ indicates the model prediction domain while $X_T$ is the \textit{top-K} similar trajectories from model's output. 
\end{enumerate}

To evaluate the model's performance, two evaluation metrics are employed:
\begin{equation}
\begin{split}
    HR@K &= \frac{1}{|\mathcal{T}|}\sum_{T \in \mathcal{T}}\frac{X_T \cap Y_T}{K}, K=N \\
    RN@K &= \frac{1}{|\mathcal{T}|}\sum_{T \in \mathcal{T}}\frac{X_T \cap Y_T}{N}, K\geq N 
\end{split}
\end{equation}
where $HR@K$ refers to the \textit{top-K} hitting ratio while $RN@K$ indicates the \textit{top-K} recall for \textit{top-N }ground truth. 
Moreover, both two similarity distance measurements (Fréchet / Hausdorff) are applied for comprehensive evaluation.  

\subsection{Trajectory Similarity Search Performance}
We compare VeTraSS's performance of representation learning with both non-learning-based methods (PCA \cite{wold1987principal}, SVD \cite{wall2003singular}, MDS \cite{kruskal1978multidimensional}) and state-of-the-art learning-based ones (SRN \cite{pei2016modeling}, NEUTRAJ \cite{yao2019computing}, node2vec \cite{grover2016node2vec}). 
\subsubsection{Quantitative Results}

The quantitative results of VeTraSS's representation learning performance are shown in Table \ref{bigtable}. 
As demonstrated, VeTraSS showcases the effectiveness over existing methods and achieves state-of-the-art performance on Porto and Geolife under two similarity measurement metrics. 
In Table \ref{bigtable}, the results in bold represent the state-of-the-art among all methods (non-learning/learning) while those in the blue background indicate the leading performance among learning-based methods.

In comparison with non-learning methods under Fréchet distance, VeTraSS demonstrates significant improvement over PCA ($+7.65\%$), SVD ($+6.74\%$), and MDS ($+3.07\%$) of HR@10 as an example on Porto.
Similar results can be observed under the Hausdorff distance: $+2.82\%$ against PCA, $+2.93\%$ against SVD and MDS.
The remarkable performance on Geolife can also be observed consistently. 

In terms of the comparison with existing learning-based models \cite{pei2016modeling, yao2019computing, grover2016node2vec}, VeTraSS consistently reaches the state-of-the-art under Fréchet / Hausdorff distance. 
Benefiting from the GNN's ability to capture global graph-level representations 
and the adaptability to different graph structures against RNN, a significant improvement can be observed (e.g. $+1.67\%$ of HR@10, $+2.26\%$ of HR@50, and $+5.75\%$ of R10@50 under Fréchet distance on Porto) when compared with NEUTRAJ \cite{yao2019computing}. Similar observations can be concluded against SRN \cite{pei2016modeling}.
This proves the effectiveness of GNN over RNN as the model backbone in the graph representation learning task.  
Since node2vec \cite{grover2016node2vec} generates representations using general graph-based data (e.g. PPI, Wiki) \cite{grover2016node2vec} instead of focusing on trajectory data specifically, it fails to precisely learn the temporal dynamics, coordinates variations, and entity movements, causing a noticeable decrease ($7.93\%$ in HR@10, $14.65\%$ in R10@50) in hitting ratio and recall.

\subsubsection{Qualitative Results}
The qualitative results of VeTraSS (on Porto as an example) are shown in Figure \ref{fig:qualitativeres}, where \textit{top-3} similar trajectories are retrieved based on the source trajectory on the left. Specifically, the results of ground truth (by computing the actual distance using the Euclidean function in the dataset) and our model are shown. 
From this figure, it is evidently to conclude that the retrieval results of VeTraSS closely match the ground truth (e.g. Rank \#2 in VeTraSS and Rank \#1 in ground truth), showcasing the effectiveness of VeTraSS in searching similar trajectories.

\begin{figure}[htb]
  \centering
  \includegraphics[width=\linewidth]{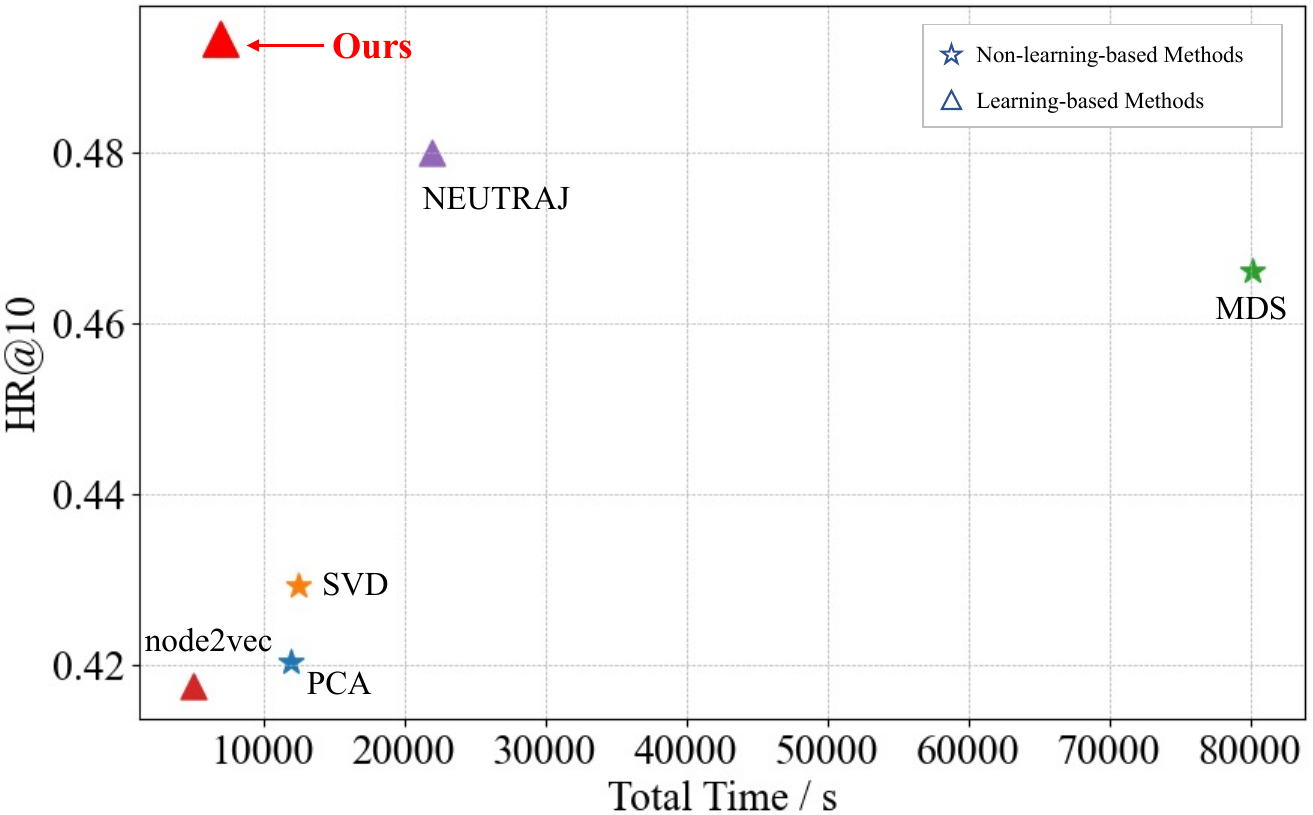}
  \caption{\textbf{Efficiency analysis.} 
  The accuracy (HR@10) against the total time for representation learning on Porto is shown, where VeTraSS (ours) represents the highest accuracy of hitting ratio while requiring one of the least time. 
  }
  \label{fig:eff}
\end{figure}

\begin{figure*}[tb]
  \centering
  \includegraphics[width=0.8\textwidth]{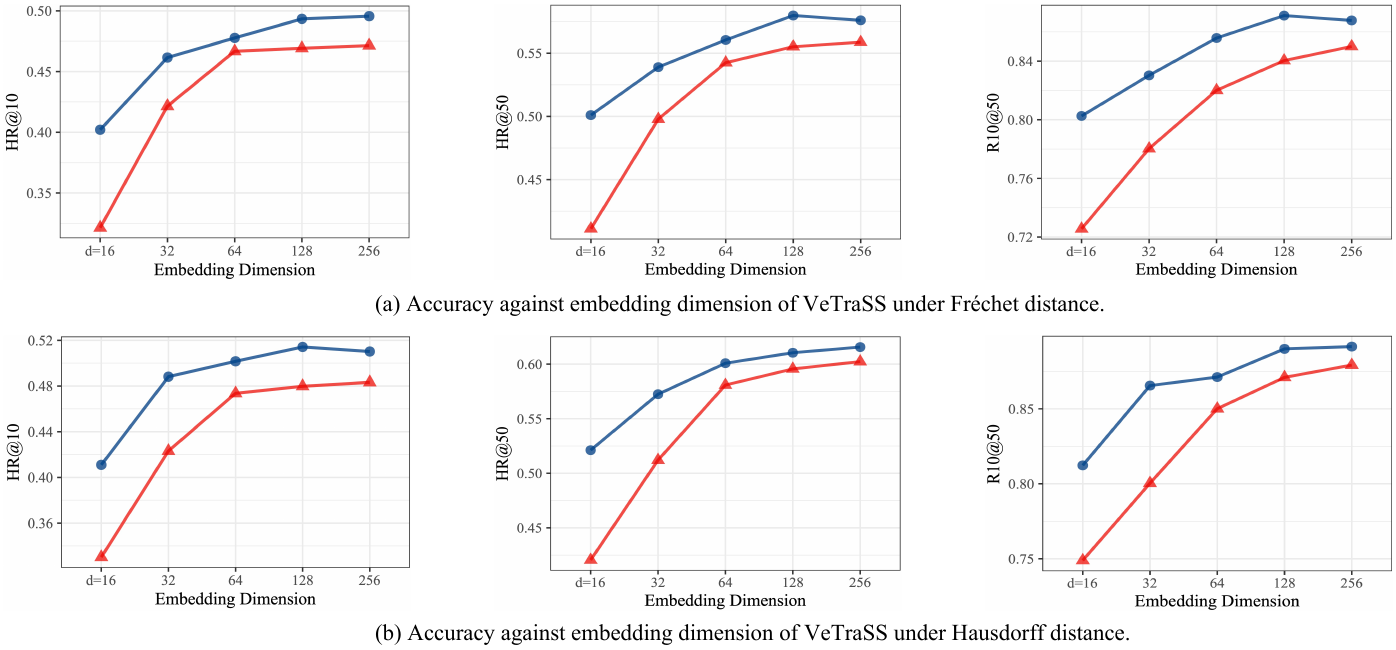}
  \caption{\textbf{Ablation studies of VeTraSS on embedding dimension on Porto.} 
  The blue line indicates the final model (VeTraSS with a multi-scale attention module) while the red line represents the single-scale version (VeTraSS$_{sing}$).
  \textbf{(a): } Accuracy against embedding dimension on Porto under Fréchet distance. 
  \textbf{(b): } Accuracy against embedding dimension on Porto under Hausdorff distance. 
  }
  \label{fig:embeddingablation}
\end{figure*}

\subsection{Efficiency Analysis}
\label{time-eff-ana}
The model efficiency analysis is shown in Figure \ref{fig:eff}, including both non-learning and learning-based approaches. 
Specifically, when compared with the
mathematical methods (PCA, SVD, MDS), VeTraSS achieves the highest accuracy (HR@10) while requiring the least time for embedding generation. 
Moreover, our model outperforms cutting-edge learning-based models significantly. We present substantial advantages in terms of both time cost and performance against NEUTRAJ \cite{yao2019computing}. In addition, we take approximately the same time cost as node2vec \cite{grover2016node2vec} while achieving considerably more accuracy than it does.

\begin{table}[]
\centering
\resizebox{1.0\linewidth}{!}{%
\begin{tabular}{@{}c|c|ccc@{}}
\toprule
\textbf{Methods}        & \multicolumn{1}{l|}{\textbf{Distance}} & \textbf{HR@10}  & \textbf{HR@50}  & \textbf{R10@50} \\ \midrule
VeTraSS$_{sing}$         & \multirow{3}{*}{Fréchet}              & 0.4692          & 0.5551          & 0.8408          \\
w/o SC                  &                                        & 0.4897          & 0.5710          & 0.8706          \\
\textbf{VeTraSS (Ours)} &                                        & \textbf{0.4968} & \textbf{0.5853} & \textbf{0.8841} \\ \midrule
VeTraSS$_{sing}$         & \multirow{3}{*}{Hausdorff}             & 0.4778          & 0.5956          & 0.8591          \\
w/o SC                  &                                        & 0.5039          & 0.5936          & 0.8752          \\
\textbf{VeTraSS (Ours)} &                                        & \textbf{0.5132} & \textbf{0.6275} & \textbf{0.8953} \\ \bottomrule
\end{tabular}
}
\caption{\textbf{Ablation study.}
This experiment quantitatively shows the effectiveness of the model's key components on the Porto dataset, including multi-scale design, sequential connection, and embedding dimension.
}
\label{ablationtable}
\end{table}

\subsection{Ablation Studies}
We conduct a comprehensive analysis of the ablation studies on the model architecture to investigate the impact of key components, including the multi-scale graph attention module, sequential connection (SC), and embedding dimension.  The primary goal is to illustrate the model's performance and the importance of these elements in embedding generation for trajectory similarity search.

\noindent
\textbf{Multi-scale Attention Module.} From Table \ref{ablationtable}, a significant drop in all the metrics can be observed when using a single-scale attention structure.
Multi-scale attention on multi-scale graphs retains more similarity information and feature associations between nodes (e.g. higher-order neighbors retrieval) than single-scale structure, and these hierarchical features can only be extracted in the higher-scale similarity graphs. Therefore, VeTraSS without such architecture fails to learn deep features for graph representation learning.

\noindent
\textbf{Sequential Connection.} Hierarchical features and representations can be learned through sequential connections to allow the model to learn complex features for graph understanding. As proven in \cite{lin2017feature, hu2020dasgil}, the fusion between multi-scale features will instruct the model to learn deep and superficial levels of abstraction, effectively enhancing the model's capabilities of graph representation learning. 

\noindent
\textbf{Embedding Dimension.} 
The ablation studies on the embedding dimension of the representation vectors are shown in Figure \ref{fig:embeddingablation}. 
When the embedding dimension increases, the performance of the two models (single/multiple attention module) shows a significant increase followed by a stable trend.
Increasing the embedding dimension will lead to an improvement in the model's ability to represent each sample, enabling the representation space to be closer to the original high-dimensional space. Therefore, a large embedding dimension ($\sim 128$) is eventually chosen in the graph representation by VeTraSS.

\section{Conclusion}
This paper introduces a novel pipeline, VeTraSS, for vehicle trajectory similarity search through graph construction and graph representation learning. 
VeTraSS first constructs the graph from the original trajectory dataset by generating a distance matrix and applying a threshold determination strategy. Each trajectory is mapped into a node in the graph, while the similarity degree is converted into edge connection weights. 
Afterward, a novel GNN based on the attention mechanism with multiple layers is designed. The final output as the embedding representation of the original data is concatenated from each layer. 
Extensive experiments on Porto and Geolife datasets strongly prove the effectiveness of VeTraSS, where our model outperforms existing work under multiple evaluation metrics and reaches the state-of-the-art. Moreover, we conduct a comprehensive analysis of the ablation studies to investigate the functionality of each key component of the model. 
Benefiting from the novel design and remarkable performance, VeTraSS showcases its potential to be applied to vehicle trajectory similarity search tasks for autonomous driving in the real world.

{
    \small
    \bibliographystyle{ieeenat_fullname}
    \bibliography{main}
}


\end{document}